\documentclass{mva_style}
\pdfoutput=1
\usepackage{graphicx}
\usepackage{xcolor}
\usepackage{svg}
\usepackage{tabularx}
\usepackage{booktabs}
\usepackage{multirow}
\usepackage{xargs}
\usepackage[normalem]{ulem}
\usepackage{url}

\finalcopy 

\begin{document}
\title{Pix2Point: Learning Outdoor 3D Using\\ Sparse Point Clouds and Optimal Transport}
\author{R. Leroy,\\ P. Trouvé-Peloux, F. Champagnat\\
ONERA, DTIS\\
F-91123 Palaiseau, France\\
{\tt\small remy.leroy@onera.fr}
\and
B. Le Saux\\
ESA/ESRIN $\Phi$-lab\\
Frascati (RM), Italy\\
{\tt\small bertrand.le.saux@esa.int}
\and
M. Carvalho\\
UPCITI\\
Montreuil, France\\
{\tt\small marcela.carvalho@upciti.com}
}

\maketitle

\begin{figure*}[!h]
\centering
\includegraphics[width=0.75\textwidth]{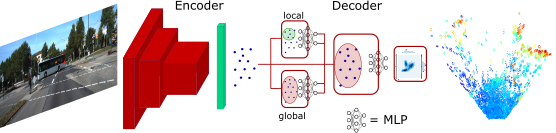}
\caption{Pix2Point: 3D point set prediction for real outdoor scenes from a single image. A first 2D CNN module encodes the RGB image for the following fully connected layer to predict a coarse 3D point cloud. Then a 3D point-wise densification module ensures to lift the number of points of this point cloud using PointNet-like MLP. All models are trained end-to-end to minimise point set distances, \eg, Optimal Transport or chamfer distances.}
\label{fig:pix2point-abstract}
\end{figure*}

\section*{\centering Abstract}
\textit{
  Good quality reconstruction and comprehension of a scene rely on 3D estimation methods. The 3D information was usually obtained from images by stereo-photogrammetry, but deep learning has recently provided us with excellent results for monocular depth estimation. Building up a sufficiently large and rich training dataset to achieve these results requires onerous processing. In this paper, we address the problem of learning outdoor 3D point cloud from monocular data using a sparse ground-truth dataset. We propose Pix2Point, a deep learning-based approach for monocular 3D point cloud prediction, able to deal with complete and challenging outdoor scenes. Our method relies on a 2D-3D hybrid neural network architecture, and a supervised end-to-end minimisation of an optimal transport divergence between point clouds. We show that, when trained on sparse point clouds, our simple promising approach achieves a better coverage of 3D outdoor scenes than efficient monocular depth methods.
}

\section{Introduction}
\label{sec:intro}
Geometry estimation is a prior requirement for autonomous agents to comprehend their environment, progress and interact within it. This fundamental task leads to various estimation methods for 3D reconstruction. For long, the parallax between images has been used~\cite{faugeras-3DCV-book1993,hartley-zisserman-MVG-book2004}, either between two cameras in stereo mode or between multiple acquisitions after displacement. Recently, deep learning techniques have revolutionised 3D estimation from images, allowing even to obtain excellent results with a single view~\cite{amiri-2019-semidepth,carvalho-18icip-losses,eigen-2014,fu-2018-dorn,lee-2019-bts}. 
These impressive results rely upon large and highly accurate databases like KITTI~\cite{Geiger2013IJRR} that involve the simultaneous collection of stereo pairs and LiDAR data, post-processing and temporal integration in order to provide accurate, reliable and dense ground-truth for learning purposes. The overall process requires large scale cooperation and is therefore lengthy. In this paper, we address the question of performing 3D estimation from monocular data with much lighter requirements in terms of training data input.
Besides, if monocular point sets prediction has been proposed in the literature, it is only for the reconstruction of a single 3D object~\cite{fan-su-guibas-2017-psgn-3d,mandikal-babu-2019-dpcr-3d,pumarola-ferrari-2020-cflow-3d}.

In contrast, we propose here \textit{(i)} one of the first approaches to reconstruct a 3D point cloud for an entire outdoor scene given only a single image using a 2D-3D hybrid neural network architecture, illustrated in Fig.~\ref{fig:pix2point-abstract}; \textit{(ii)} an end-to-end learning scheme of the hybrid model using a sparse point clouds dataset; \textit{(iii)} a cost function from the optimal transport (OT) that makes possible to obtain good coverage of the scenes on the KITTI dataset~\cite{Geiger2013IJRR} 
\textit{(iv)} better performances than state-of-the-art monocular depth map prediction methods when trained with sparse data. This illustrates the advantage of the direct prediction of point clouds, in the case of a low-density ground-truth 3D dataset.

\section{Related Work}
\label{sec:related}
The monocular 3D estimation task has been addressed in terms of depth maps prediction with the following works~\cite{bhat-2020-adabins,carvalho-2018-dfd,eigen-2014,fu-2018-dorn,lee-2019-bts,zhu-2020-monocular-depth-continuous-3d-loss,saxena-2005-learning-depth} but as already mentioned, they are trained with large and rich datasets. Recently, 3D estimation has also been addressed in terms of 3D point set with PSGN by Fan \etal~\cite{fan-su-guibas-2017-psgn-3d}, a method that aims to predict the envelope of an object as an unordered set of points using a single view of it and its location in the image. Mandikal and Babu~\cite{mandikal-babu-2019-dpcr-3d} address the limitations of PSGN regarding the poor number of predicted points with DensePCR, a pyramidal structure allowing to enlarge the number of points. Xia \etal~\cite{xia-2018-realpoint3d}, also tackles the generation of a monocular point cloud for objects using prior knowledge over their shapes, making it robust to occlusions and varying poses. A generative flow-based model allowing single view object point cloud prediction has recently been proposed with C-flow~\cite{pumarola-ferrari-2020-cflow-3d}. It leverages a back-and-forth prediction loop from image to point cloud, then to image for consistency. \par 
The aforementioned point cloud works only consider the reconstruction of a single 3D object model, \ie, on data that are obtained through demanding procedures, either scanned objects using RGB-D sensors or laser scanners, or handcrafted models. These procedures do not apply to real-life outdoor scenes with various settings and where lies multiple objects. 
\par\cite{denninger-2020-3d-viewport} tackles the problem of monocular volumetric reconstruction with occlusion completion for indoor scenes. In addition to the input RGB image, this approach requires a corresponding normal image that is hardly obtainable for outdoor scenes.
To our knowledge, we are the first to propose a deep learning method for the reconstruction of complex outdoor scenes in the form of 3D point cloud, solely conditioned by a single RGB image, and learned on sparse point clouds.
Our technical contributions with respect to previous works are: a performance study of the 2D-3D hybrid neural network architecture for point cloud reconstruction of scenes "in-the-wild", end-to-end training with OT loss optimisation on sparse point clouds up to ten thousands elements. Finally, we compare our point set prediction approach to state of the art depth map prediction approaches, when they are all trained on the same sparse dataset. 

\section{Method}
\label{sec:method}
Given a single colour image, our method, namely Pix2Point, predicts a set of 3D point coordinates with an arbitrary number of elements fixed before training. Fig.~\ref{fig:pix2point-abstract} shows an overview of the proposed method. Like DensePCR, our architecture consists of an encoding-decoding module to predict a first coarse point cloud that will be enriched using a densification module. Our model, unlike DensePCR, is trained to minimise an optimal transport divergence over the point coordinates in an end-to-end fashion.\par
\textbf{Encoding}
\label{subsec:encoding}
The encoding block is a series of convolution, pooling and normalisation layers to extract a feature description of the full RGB input image, which is then processed by a fully connected layer to obtain a first coarse set of 3D point coordinates. We explore and compare several encoding approaches following either VGG, DenseNet and ResNet architectures in section \ref{sec:results}. We refer to them as \textit{backbones}. \par
\textbf{Decoding}
\label{subsec:decoding}
We refer to decoding as the densification of the first coarse 3D point cloud. 
We duplicate every point $n$ times, and to describe each point we concatenate: the 3D coordinates, both global point cloud feature vector and the corresponding local feature vector, obtained using dedicated PointNet-like shared multi-layered perceptron (MLP)~\cite{qi-su-guibas-2017-pointnet,qi-su-guibas-2017-pointnet++}, and lastly a grid alignment feature vector in order to identify every clone of the same point and to suggest geometric information between every clone. This point description is processed by another shared MLP resulting in 3D coordinates for 1 point.
\par
\textbf{Optimisation}
The performance of our approach mainly depends on the training loss function. Unlike the depth map prediction methods which exploit the gridded structure of the image for the evaluation of errors, our method uses distances between unordered point sets. In particular, it requires an additional, computationally expensive step to match points from the predicted and the target point clouds. We expose two usual distances for this task.

\textbf{Chamfer distance} 
The chamfer distance is the average of squared Euclidean distances to the nearest neighbour from one set to the other. 

\textbf{Optimal transport or OT distance}
 To compute this distance, also known as Earth Mover's distance, one has to find a one-to-one mapping from one set to the other that minimises the sum over each point of the squared distance between them and their corresponding image. This minimised sum is called the OT distance and it informs about the eventual discrepancy between point sets distributions.

The exact computation of an OT distance is time and memory expensive especially for several thousand elements, hence, we consider in our work an approximation of the OT distance obtained by adding a regularising term and solving the Sinkhorn-Knopp algorithm~\cite{cuturi-2013, feydy-2019-sinkhorn}. 

\section{Experimental Settings}
\label{sec:experiment-setting}
This section presents the dataset we are considering for real scene point cloud estimation from a single image and defines several criteria for reconstruction performance evaluation.

\subsection{Dataset}
\label{subsec:dataset}
To assess our method, we operate on RGB image sequences of real urban scenes and corresponding LiDAR point cloud acquisitions from the KITTI depth estimation benchmark dataset~\cite{Geiger2013IJRR}. Every scene point cloud is an accumulation of filtered LiDAR acquisitions over few successive time instants. We use the split defined by Eigen \etal~\cite{eigen-2014}, that is 22 600 training scenes and 697 testing scenes. 

\subsection{Evaluation Metrics}
\label{subsec:metrics}
There is no reference value or method for the KITTI scene reconstruction task regarding 3D point clouds. Therefore, we propose to measure performances using the completeness and accuracy criteria from~\cite{Tylecek2018rms}, and also propose the relative accuracy. All 3 measures are defined as follows:

\par \textbf{Completeness} is the coverage in per cent of the target point cloud by the predicted points. A target point is covered if a predicted point lies in its surrounding (\ie fixed radius ball). We evaluate completeness values for radius of $ 50 $ cm, $25$ cm and $ 10 $ cm.
\par \textbf{Accuracy} is the distance $d$, in meter, from the $r$-th percentile of the distances to the nearest neighbour, from the predicted point cloud to the ground-truth point cloud. It measures the greatest distance to the nearest neighbour among the predicted points closest to the ground truth. We choose $r<90\%$ to include most of the points and discard eventual outliers.
\par \textbf{Relative accuracy} is similar to the accuracy, where every distance to their nearest neighbour is divided by the norm of the corresponding target point. It provides a higher penalty to short-range predictions.

\subsection{Implementation Details}
\label{subsec:implement}
Experiments were conducted using the \textit{Pytorch} Framework~\cite{pytorch-2019}. We kept the original image resolution and cropped every picture to $1224 \times 370$ pixel definition. Due to heavy computational cost for loss back-propagation, parameters were updated after every sample forward, making batch normalisation ineffective. Instance normalisation was applied instead~\cite{ulyanov-2016-instance-normalization}. The number of predicted points was determined to fully load the 8GB GPU during training. Therefore, $2500$ elements point clouds are first predicted by the fully connected encoding module, that are then up-scaled by a DensePCR-like module making a point cloud with 10k elements.
\par
\textbf{Training vs. Testing point clouds:}  Using an OT loss enables, in principle, the comparison of any ground-truth point cloud to the predicted one, however, in practice, computation and optimisation of an OT loss are computationally much more efficient with point sets of equal cardinality. Therefore, ground-truth point cloud databases are randomly subsampled to 10k points, that is as many points as Pix2Point predicts.
When testing we measure performances to the whole ground-truth point cloud.

\section{Experimental Results}
\label{sec:results}
In this section we first present the performances of Pix2Point using various encoding backbones and losses, then we compare our method to depth map prediction approaches through evaluation metrics defined in~\ref{subsec:metrics}.



\subsection{Network parameter study}
\label{subsec:abaltion}

We trained several models with varying encoding backbones and loss functions. We considered the following configurations: Pix2Point architecture with VGG backbone and training on the minimisation end-to-end of either the chamfer or OT distance, and Pix2Point with ResNet backbone and minimising the OT distance. The performances of these models are given in Table~\ref{tab:perf} respectively as P2P-VGG-C, P2P-VGG-OT and P2P-ResNet-OT. From these figures, we can notice that the minimisation of chamfer distance thrives toward predictions with low local error, and minimising the OT distance grants predictions with higher completeness, hence, better coverage of the scenes. In order to find if these distances could help each other, combinations of both distances have been tested. However, they lead to convergence issues during training and overall worst performances due to opposite objectives of the distances.
Changing the backbone from VGG to ResNet has also a slight impact on the completeness and accuracy. The small gain in relative accuracy indicates that far predictions are more accurate.
We also provide a comparison to a similar image-to-point-cloud approach, DensePCR~\cite{mandikal-babu-2019-dpcr-3d}, initially proposed for 3D graphics models.


\subsection{Comparison to depth prediction approaches}
\label{subsec:dense_comparison}
\renewcommand{\arraystretch}{0}
\begin{figure*}[ht]
    \centering
    
    \begin{tabular}{*{6}{@{}c@{}}}
        \hfill\includegraphics[width=0.14\textwidth]{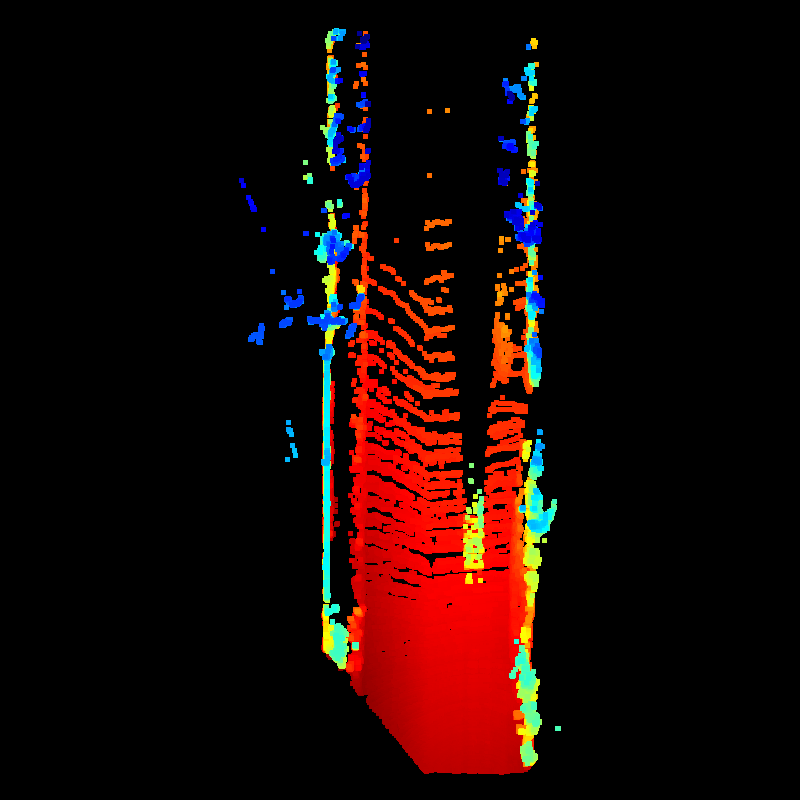} &
        
        \includegraphics[width=0.14\textwidth]{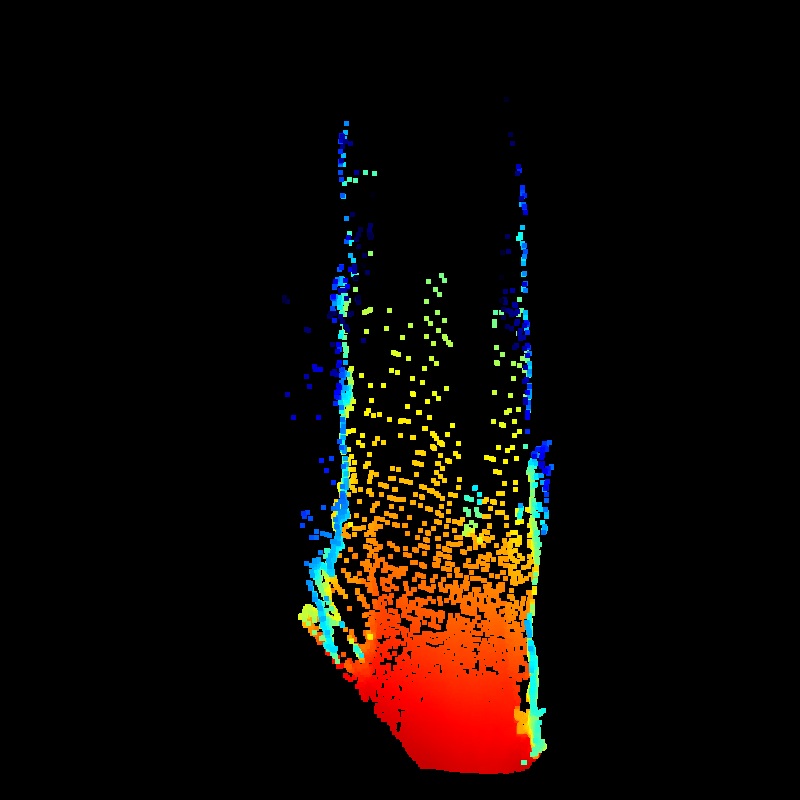} &
        \includegraphics[width=0.14\textwidth]{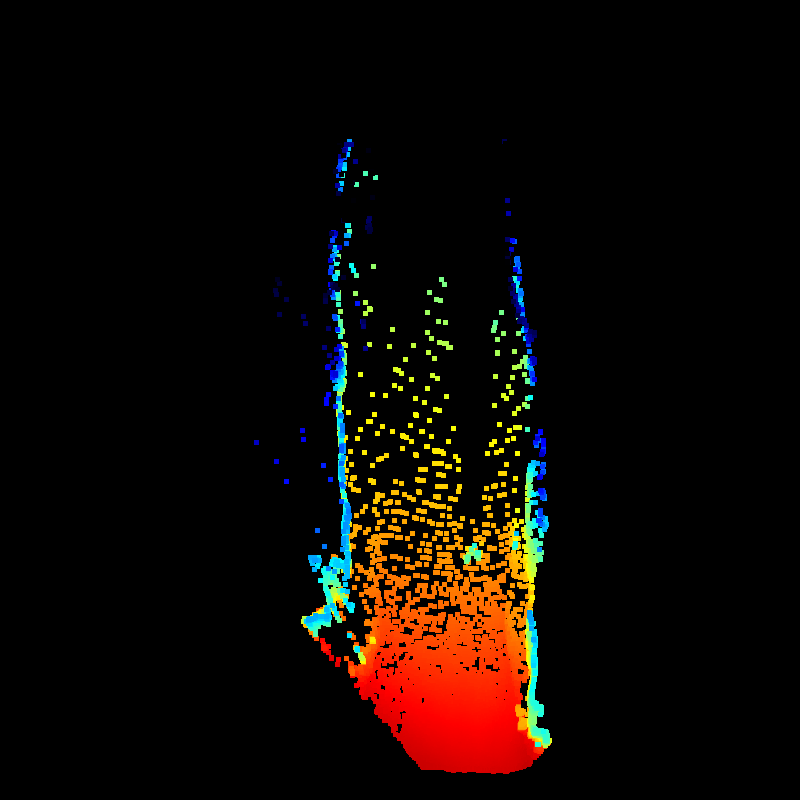} &
        \includegraphics[width=0.14\textwidth]{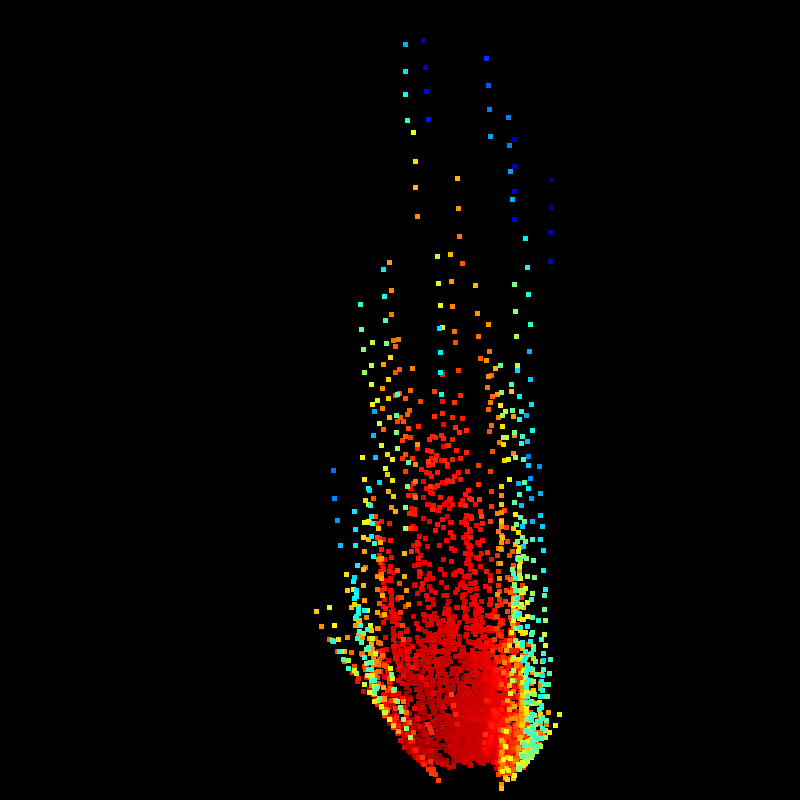} &
        \includegraphics[width=0.14\textwidth]{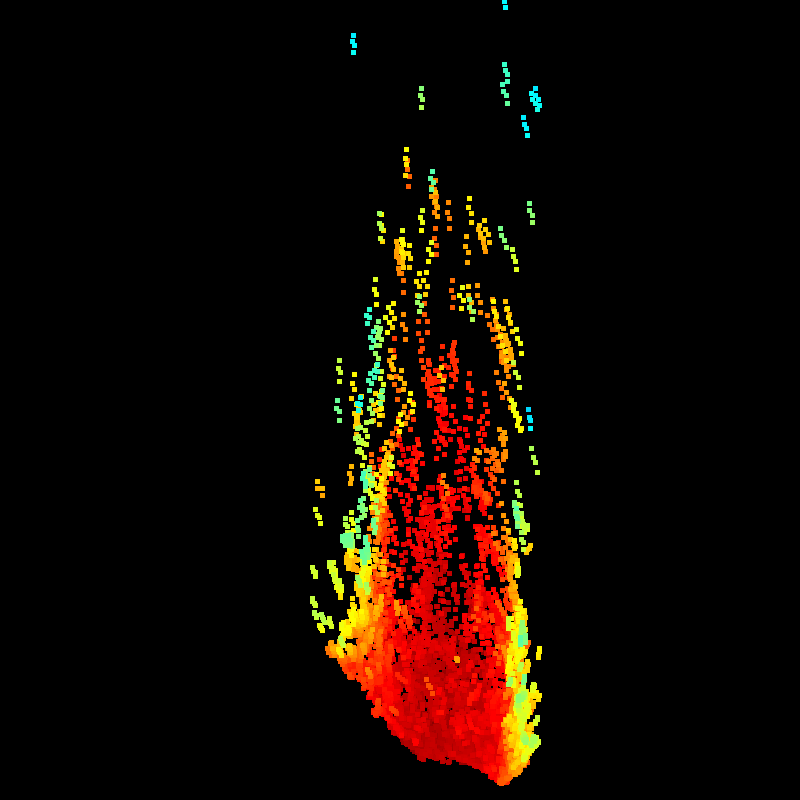} &
        \includegraphics[width=0.14\textwidth]{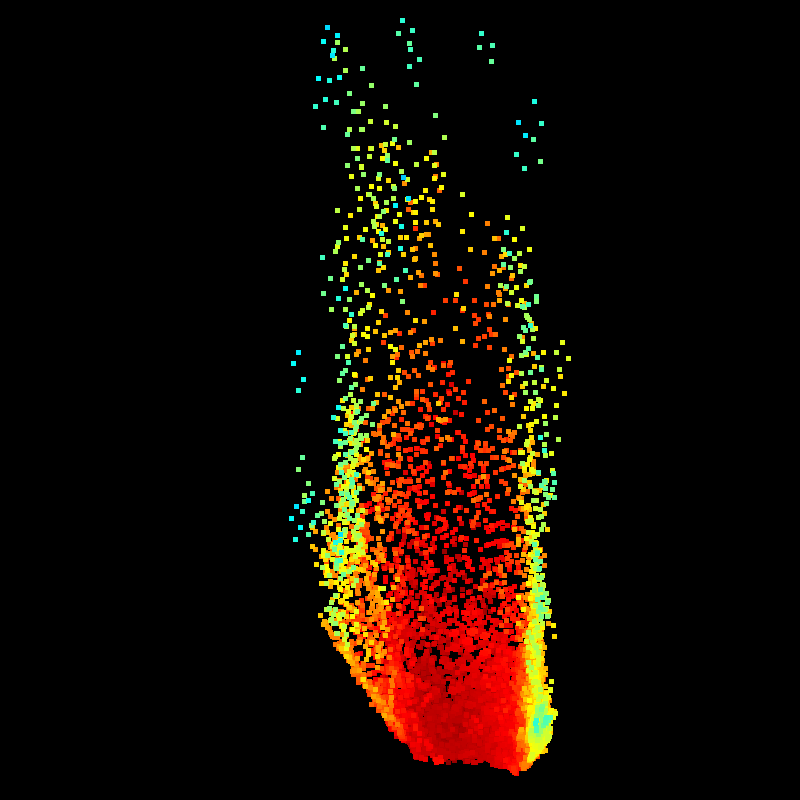} 
         \\
        \includegraphics[width=0.25\textwidth]{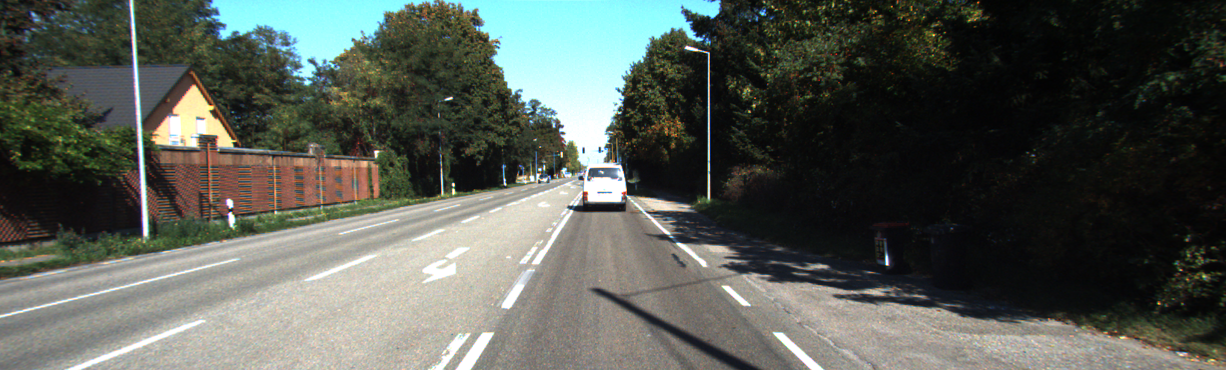} &
        
        \includegraphics[width=0.14\textwidth]{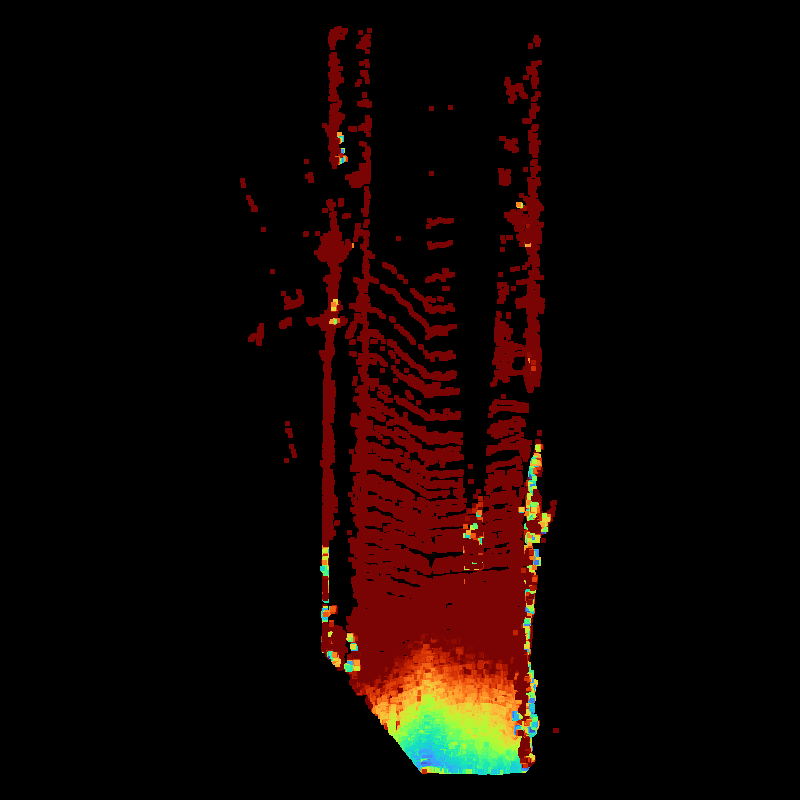} &
        \includegraphics[width=0.14\textwidth]{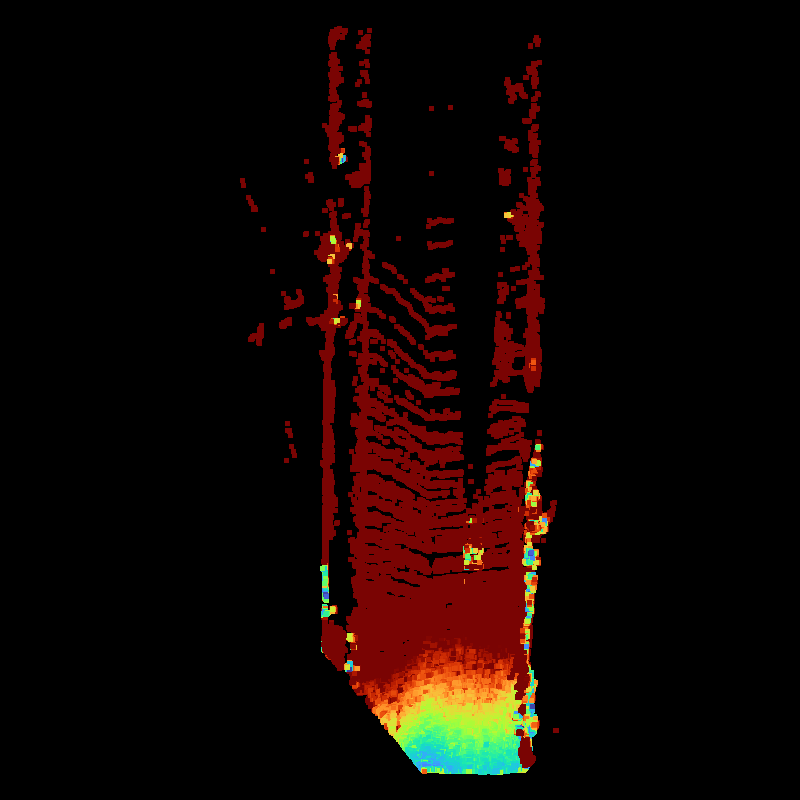} &
        \includegraphics[width=0.14\textwidth]{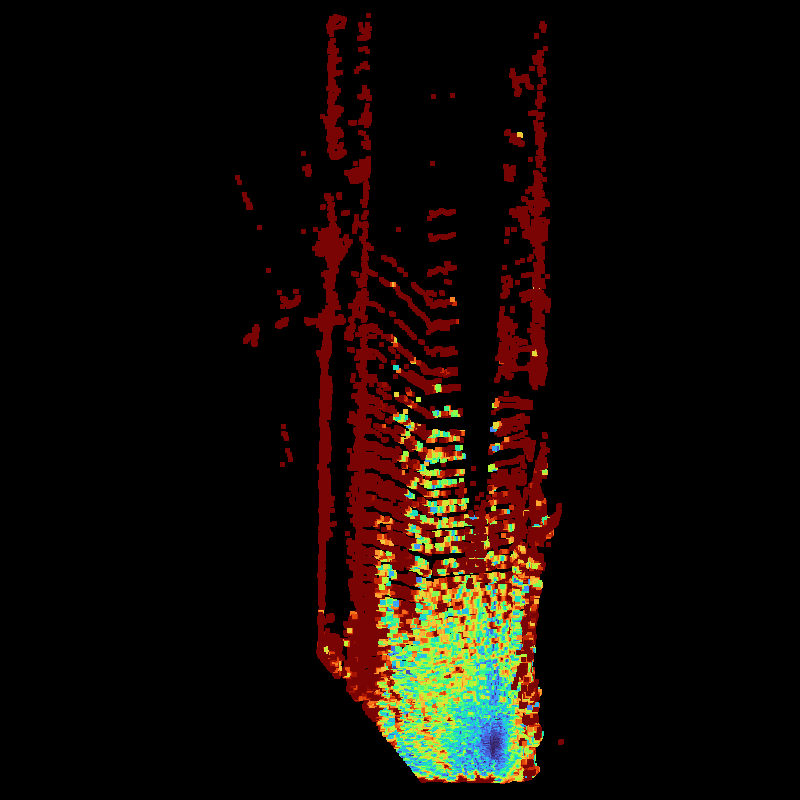} &
        \includegraphics[width=0.14\textwidth]{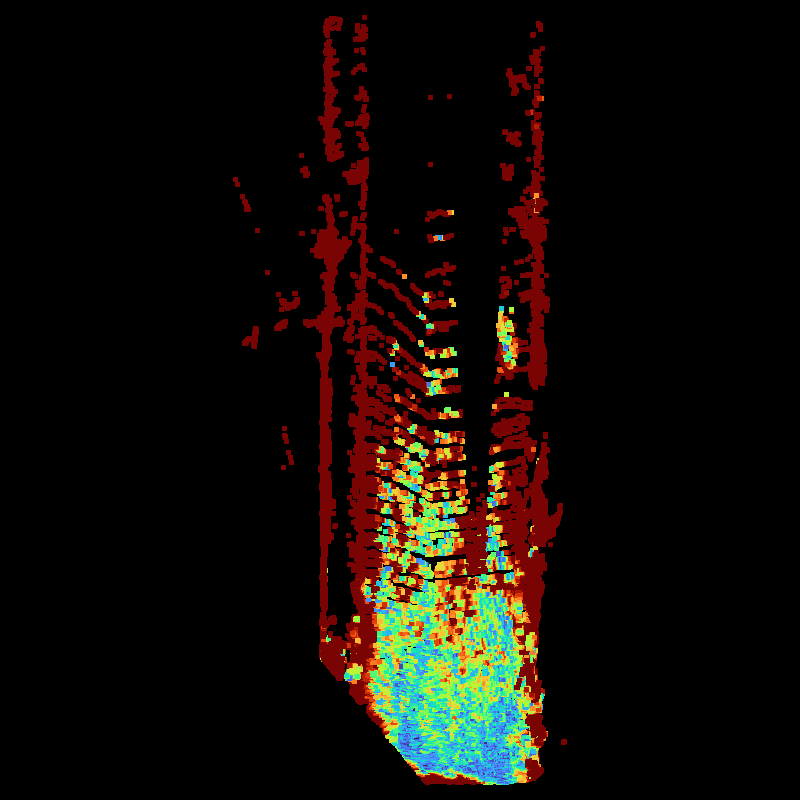} &
        \includegraphics[width=0.14\textwidth]{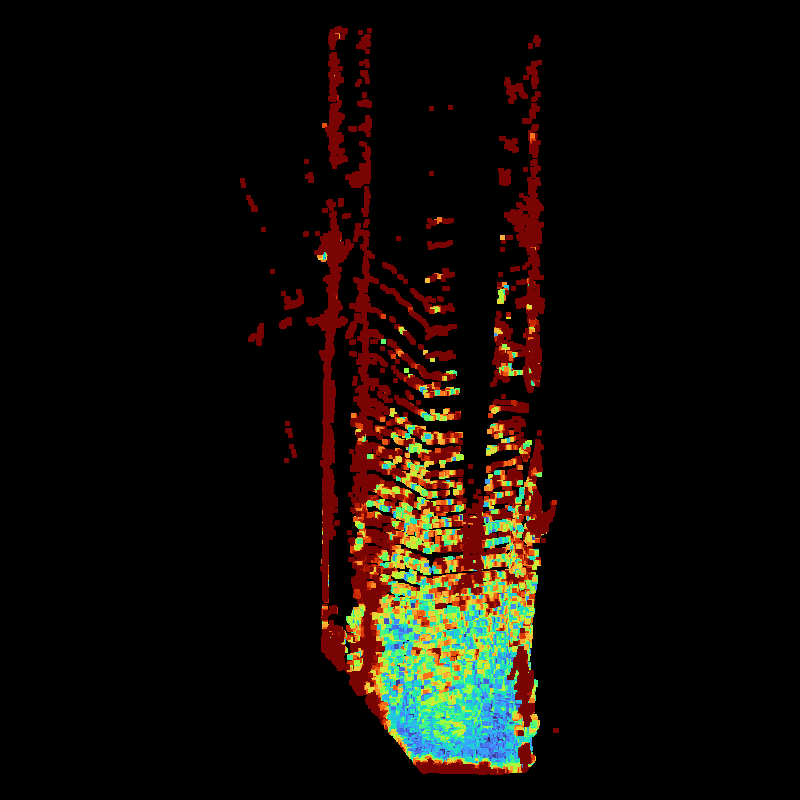} 
        \\
        \vspace{0.2em}
         \\
        \hfill\includegraphics[width=0.14\textwidth]{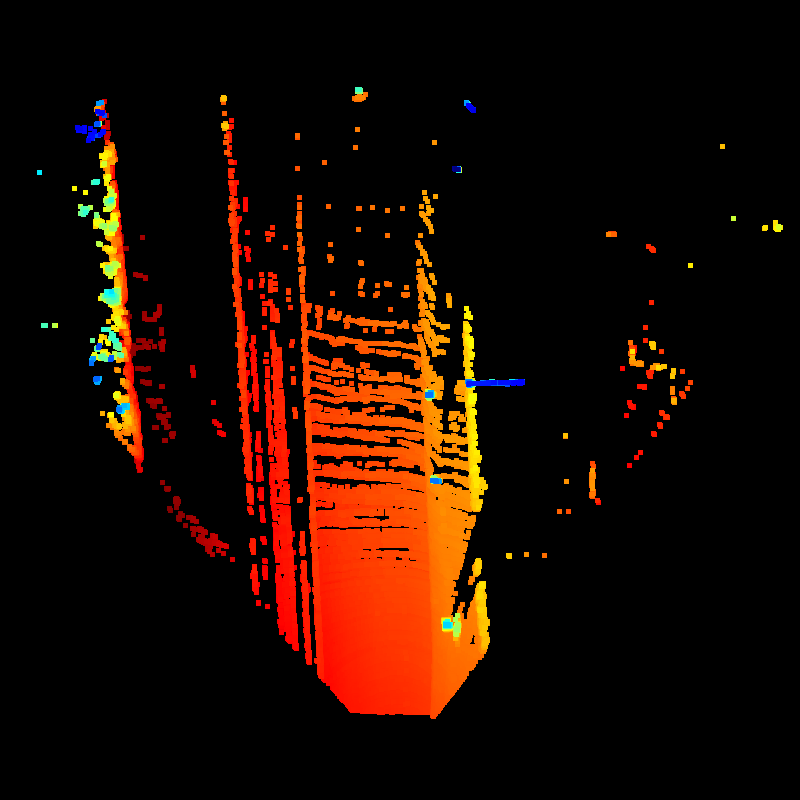} &
        
        \includegraphics[width=0.14\textwidth]{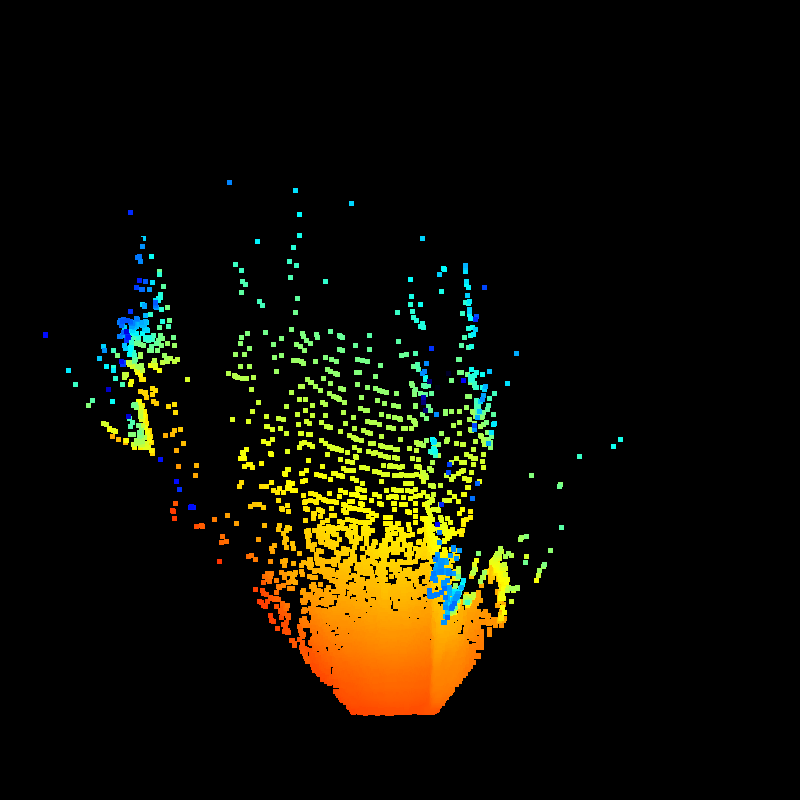} &
        \includegraphics[width=0.14\textwidth]{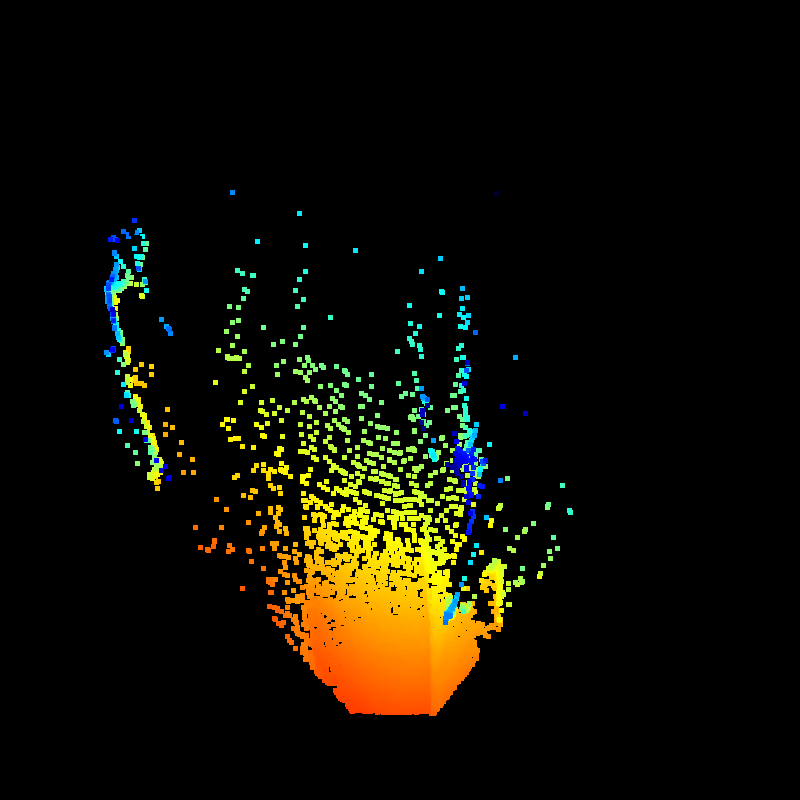} &
        \includegraphics[width=0.14\textwidth]{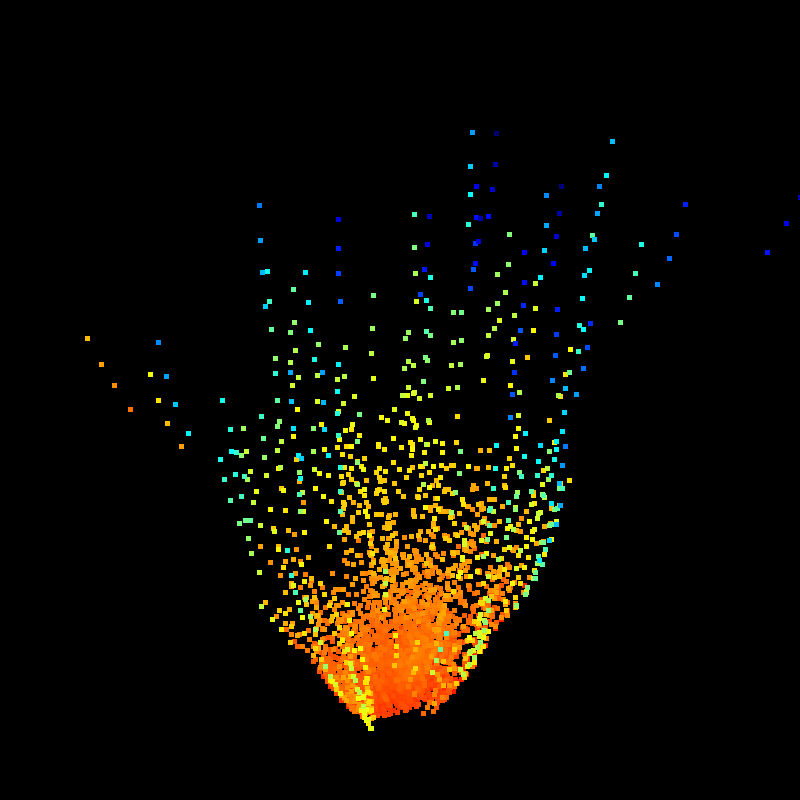} &
        \includegraphics[width=0.14\textwidth]{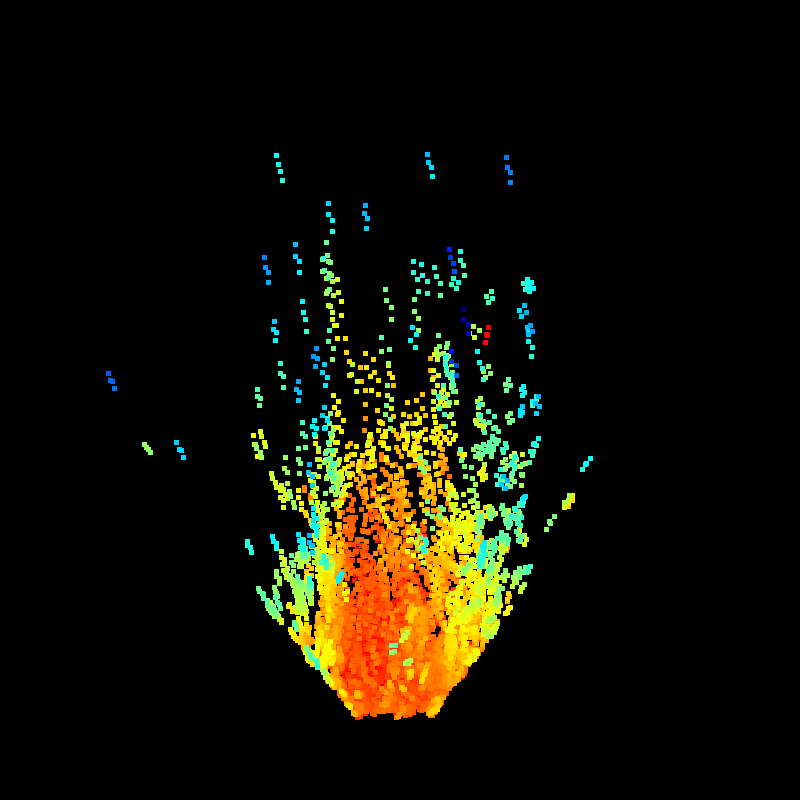} &
        \includegraphics[width=0.14\textwidth]{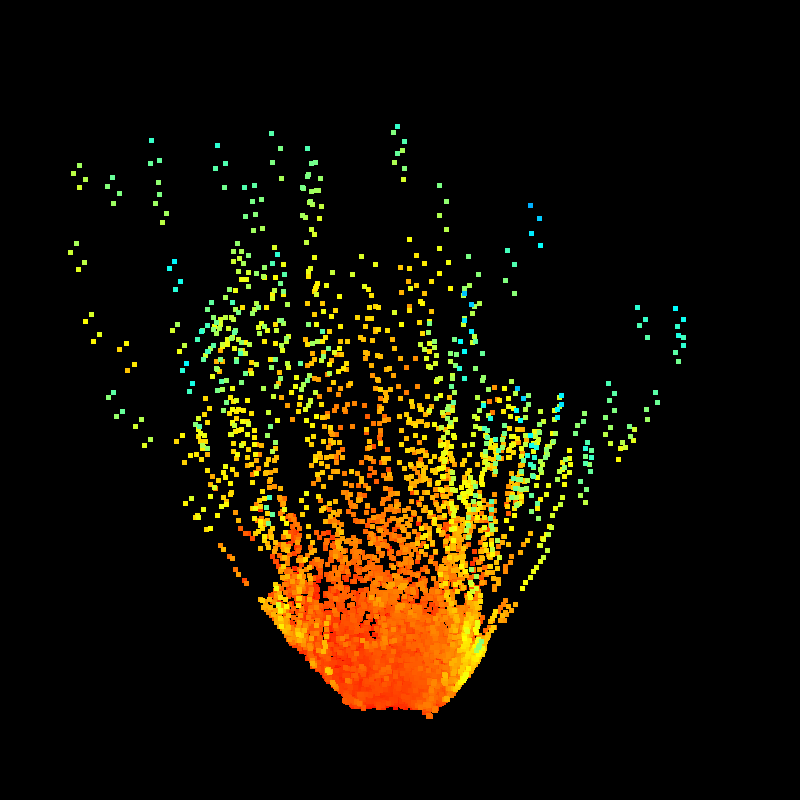} 
         \\
        \includegraphics[width=0.25\textwidth]{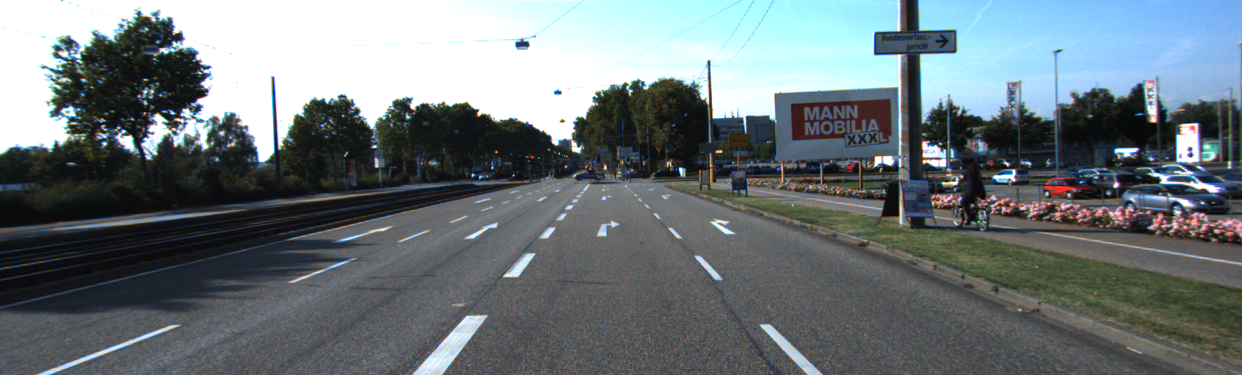} &
        
        \includegraphics[width=0.14\textwidth]{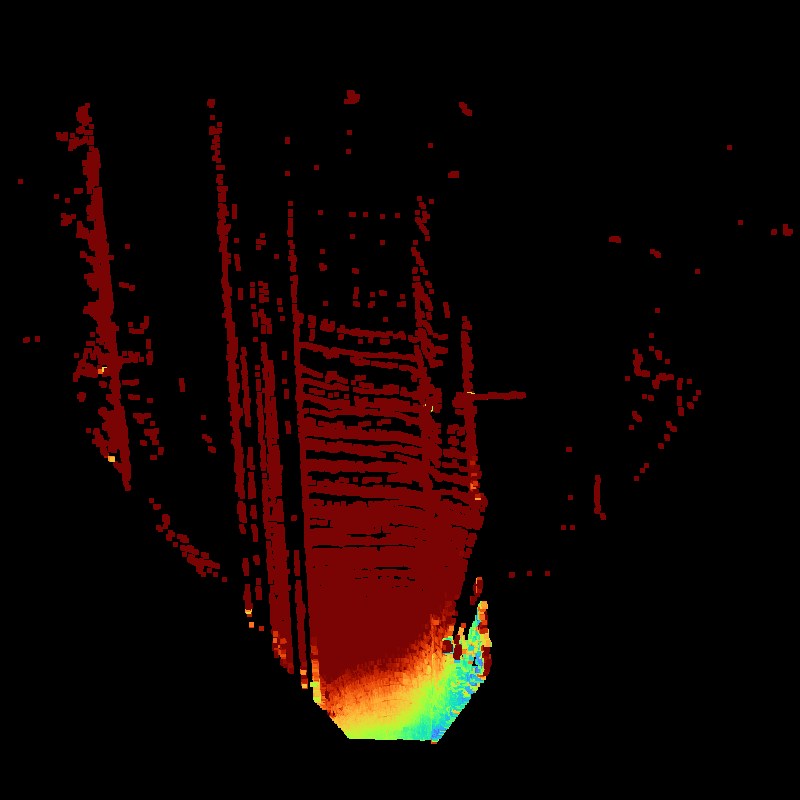} &
        \includegraphics[width=0.14\textwidth]{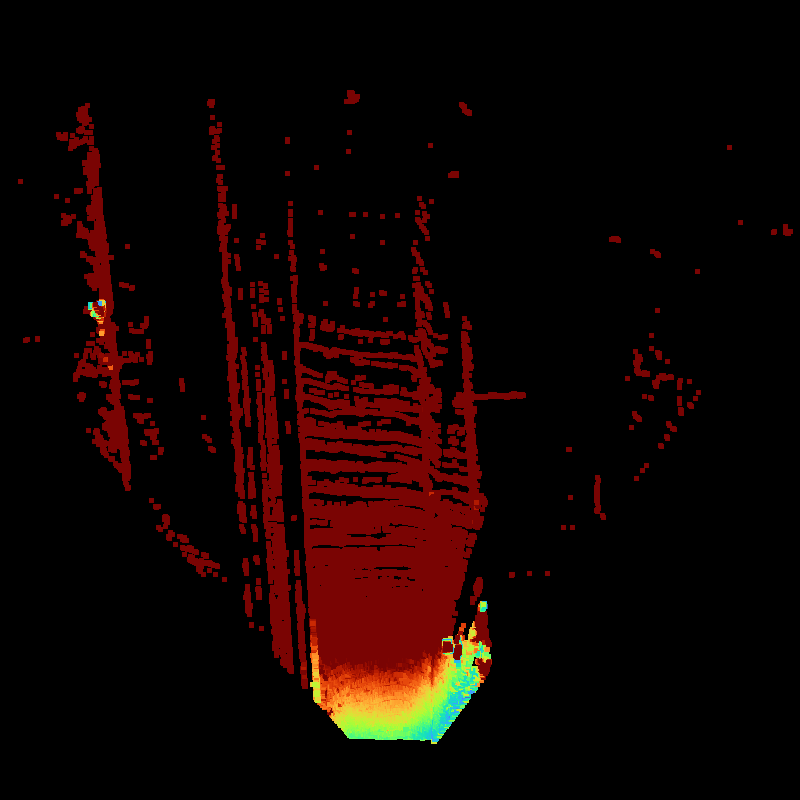} &
        \includegraphics[width=0.14\textwidth]{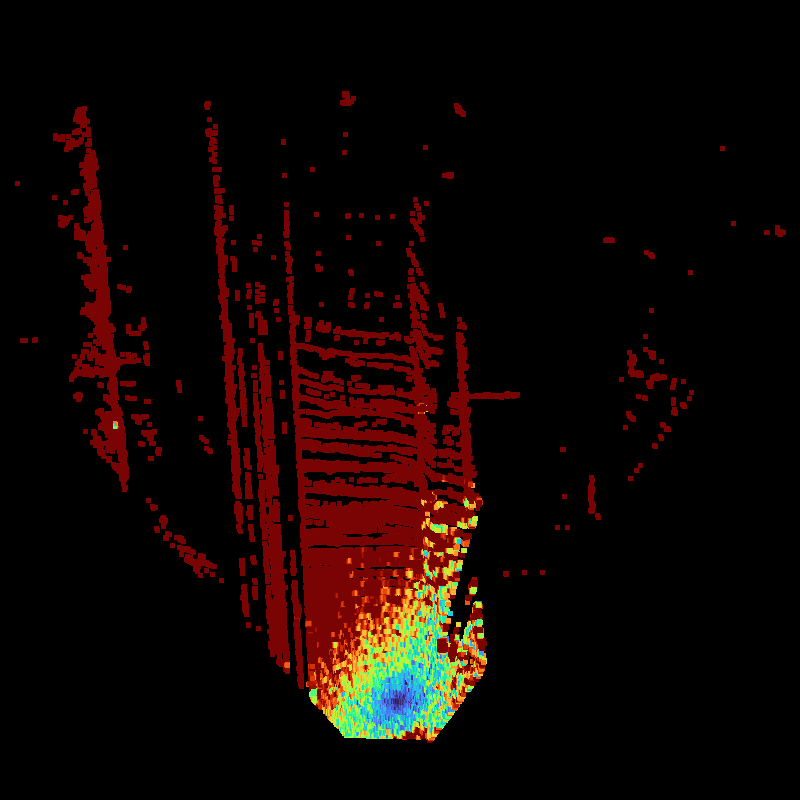} &
        \includegraphics[width=0.14\textwidth]{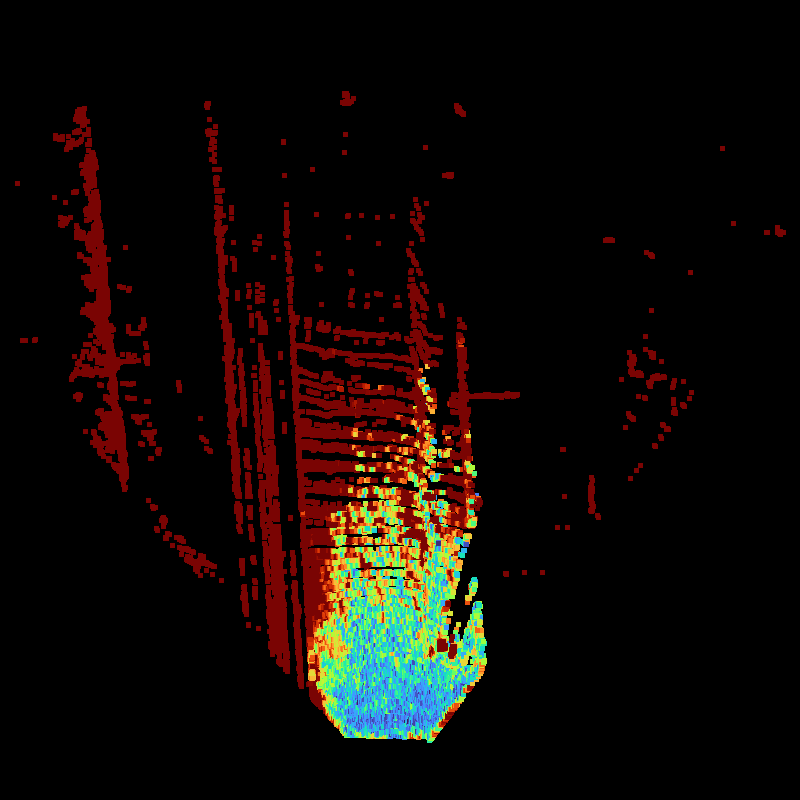} &
        \includegraphics[width=0.14\textwidth]{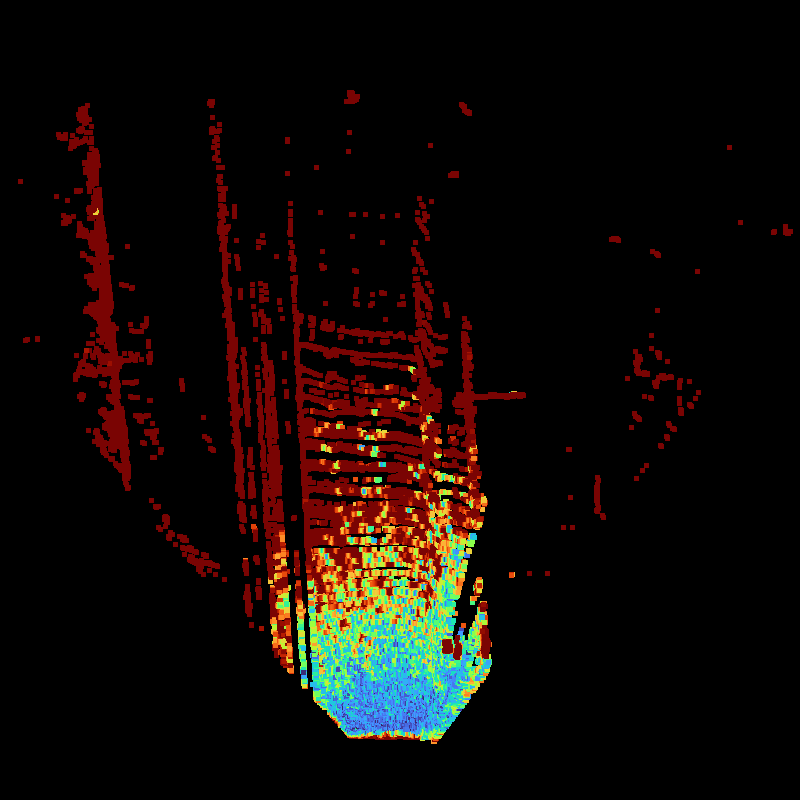}
        \\
        \vspace{0.5em}\\
        \scriptsize RGB and ground-truth scene
        & \scriptsize AdaBins &\scriptsize BTS&\scriptsize P2P-VGG-C &\scriptsize P2P-VGG-OT &\scriptsize P2P-ResNet-OT\\
        \vspace{1em}

    \end{tabular}
    \caption{For each scene, first row: 3D ground-truth and predictions for the RGB image according to AdaBins~\cite{bhat-2020-adabins}, BTS~\cite{lee-2019-bts}, our Pix2Point VGG-chamfer, VGG-OT and ResNet-OT, all trained on 10k points. We follow the 3D representation of~\cite{caccia-courville-pineau-2019-generative-lidar}. Bird's eye view where the colour encodes the altitude.
    Second row: the input RGB image and the ground-truth-to-prediction error map for each method. Errors are from 0 (blue) to 50cm (red).}
    \label{fig:depth_map_lidar_pred}
\end{figure*}

The current dominant approach to 3D estimation from a single image consists of predicting corresponding depth maps by leveraging the power of image-to-image translation networks. On KITTI, these methods are trained on pseudo-dense depth maps built by accumulating several consecutive LiDAR acquisitions. For comparison in a similar setting, we train two state-of-the-art models for monocular depth estimation from the KITTI challenge\footnote{\url{http://www.cvlibs.net/datasets/kitti/eval_depth.php?benchmark=depth_prediction}}, namely BTS~\cite{lee-2019-bts} and AdaBins~\cite{bhat-2020-adabins}, on the same 10k-point-cloud as Pix2Point. At inference time, dense depth maps are projected back to 3D using known camera parameters. Performance comparison with various flavours of Pix2Point is reported in Table~\ref{tab:perf}.\par

These results show that Pix2Point, with only 15M parameters, trained with Chamfer distance performs better than BTS and Adabins, respectively 45M and 78M parameters. Moreover, when trained with the OT distance, Pix2Point accuracy decreases but it covers three times more points than depth map approaches for the closest neighbourhoods. 
These observations can be made through Figure~\ref{fig:depth_map_lidar_pred} where we show point cloud predictions and the coverage error map for each method (for comparison all predictions are visualized with 10k points). This error map displays for each ground-truth point the distance to its nearest predicted point. We display the error from 0 to 50cm using the jet colourmap. While AdaBins and BTS preserve fine features, all Pix2Point variants achieve better coverage of the scene and a lower error, especially for far-away elements and small size features, that are not retrieved by AdaBins and BTS (see for instance the right part of the bottom scene).

\renewcommand{\arraystretch}{0.88}
\begin{table}[ht]
    \caption{Comparison of 3D scene reconstructions on KITTI. We report completeness and accuracy. All methods are trained with 10k point clouds.}
    \label{tab:perf}
    \centering
    {\small
    \begin{tabular}{l|c c c | c c}
    \toprule
        Approaches & \multicolumn{3}{c|}{Completeness $\uparrow$(in \%)} & \multicolumn{2}{c}{Accuracy $\downarrow$}\\
                & 50cm & 25cm & 10cm & \multicolumn{1}{c|}{in m} & rel.\\ \bottomrule        
        \hline
        P2P-ResNet-OT & \textbf{71.35} & \textbf{48.82} & \textbf{15.12} & 1.92 & 0.18\\
        P2P-VGG-OT & 67.4 & 47.7 & 14.7 & 1.79& 0.19\\
        P2P-VGG-C & 64.4 & 36.0 & 8.0 & \textbf{0.85} & \textbf{0.05}\\
        \midrule
        DensePCR & 59.9 & 23.5 & 3.5 & \textit{1.77} & 0.18\\
        \midrule
        BTS & 67.59 & 31.29 & 6.28 & 1.23 & 0.06
        \\
        AdaBins & 65.86 & 27.52 & 5.71 & 1.25 & 0.06\\
    \bottomrule
    \end{tabular}
    }
\end{table}

\section{Conclusion}
We proposed in this work an innovative approach to tackle point clouds reconstruction for complex outdoor scenes from a unique RGB image, using a light 2D-3D hybrid neural network. It recovers properly distributed point clouds by taking advantage of an optimal transport loss. We also provided the first benchmark for this novel task on the KITTI dataset and introduced performance metrics to assess the quality of point cloud reconstruction. We show that our method outperforms state of the art depth map prediction methods, when trained with sparse data, illustrating the interest of direct image to 3D point-cloud translation. 

\bibliographystyle{plain}
\bibliography{pix2pointMVA_camera_ready}

\end{document}